\documentclass[journal]{IEEEtran}
\usepackage{amsmath,amsfonts}
\usepackage{algorithmic}
\usepackage{algorithm}
\usepackage{gensymb}
\usepackage{array}
\usepackage[caption=false,font=normalsize,labelfont=sf,textfont=sf]{subfig}
\usepackage{textcomp}
\usepackage{stfloats}
\usepackage{url}
\usepackage{verbatim}
\usepackage{graphicx}
\usepackage{diagbox}
\usepackage{cite}
\usepackage{booktabs}
\usepackage{multirow}
\usepackage[colorlinks=true,linkcolor=blue]{hyperref}

\usepackage{enumitem}
\setlist[itemize]{leftmargin=10pt} 

\begin{document}

%
%
\title{Spatial-Temporal-Spectral Unified Modeling \\ for Remote Sensing Dense Prediction}

\author{Sijie Zhao, Feng Liu, Enzhuo Zhang, Yiqing Guo, \\ Pengfeng Xiao, \textit{Senior Member, IEEE}, Lei Bai, Xueliang Zhang*, \textit{Senior Member, IEEE},  Hao Chen*
\thanks{This work was done during the internship of Sijie Zhao at Shanghai Artificial Intelligence Laboratory. This research is supported by the the Shanghai Municipal Scienc7e and Technology Major Project. This work is partially supported by the National Natural Science Foundation of China (Grant No. 42471410), the AI and AI for Science Project of Nanjing University (Grant No. 020914380171), and by the Joint Fund for Meteorology of the National Natural Science Foundation of China (Grant No. U244220069). \textit{Corresponding author: Xueliang Zhang and Hao Chen}.}
\thanks{Sijie Zhao, Enzhuo Zhang, Xueliang Zhang, and Pengfeng Xiao are with the Jiangsu Provincial Key Laboratory for Advanced Remote Sensing and Geographic Information Technology, Key Laboratory for Land Satellite Remote Sensing Applications of Ministry of Natural Resources, School of Geography and Ocean Science, Nanjing University, Nanjing 210023, China (e-mail: zsj@smail.nju.edu.cn; Zenzhuo@smail.nju.edu.cn; yiqingguo@smail.nju.edu.cn; zxl@nju.edu.cn; xiaopf@nju.edu.cn).}

\thanks{Feng Liu is with the School of Electronic Information and Electrical Engineering, Shanghai Jiao Tong University, Shanghai 200240, China, and also with Shanghai Artificial Intelligence Laboratory, Shanghai 200232, China (e-mail: liufeng2317@sjtu.edu.cn)}
\thanks{HaoChen and Lei Bai are with the Shanghai Artificial Intelligence Laboratory, Shanghai 200232, China (email: justchenhao@buaa.edu.cn; bailei@pjlab.org.cn).}
\thanks{Codes are available at \textit{https://github.com/walking-shadow/Official\_TSSUN}}
}




\maketitle

%
%
\begin{abstract}

The proliferation of multi-source remote sensing data has propelled the development of deep learning for dense prediction, yet significant challenges in data and task unification persist. Current deep learning architectures for remote sensing are fundamentally rigid. They are engineered for fixed input-output configurations, restricting their adaptability to the heterogeneous spatial, temporal, and spectral dimensions inherent in real-world data. Furthermore, these models neglect the intrinsic correlations among semantic segmentation, binary change detection, and semantic change detection, necessitating the development of distinct models or task-specific decoders. This paradigm is also constrained to a predefined set of output semantic classes, where any change to the classes requires costly retraining. To overcome these limitations, we introduce the Spatial-Temporal-Spectral Unified Network (STSUN) for unified modeling. STSUN can adapt to input and output data with arbitrary spatial sizes, temporal lengths, and spectral bands by leveraging their metadata for a unified representation. Moreover, STSUN unifies disparate dense prediction tasks within a single architecture by conditioning the model on trainable task embeddings. Similarly, STSUN facilitates flexible prediction across multiple set of semantic categories by integrating trainable category embeddings as metadata. Extensive experiments on multiple datasets with diverse Spatial-Temporal-Spectral configurations in multiple scenarios demonstrate that a single STSUN model effectively adapts to heterogeneous inputs and outputs, unifying various dense prediction tasks and diverse semantic class predictions. The proposed approach consistently achieves state-of-the-art performance, highlighting its robustness and generalizability for complex remote sensing applications.

\end{abstract}


%
%
\section{INTRODUCTION}
\label{sec:intro}

    With the continuous advancement of remote sensing technologies and the increasing diversity of data acquisition methods \cite{intro1}, the field of remote sensing has entered a phase of rapid development \cite{intro2}. Massive and multi-source remote sensing data have been widely applied to various dense prediction tasks, playing a crucial role in applications such as urban expansion monitoring \cite{pami_building_monitor}, land cover classification \cite{pami_lulc}, disaster damage assessment \cite{intro5}, crop classification and yield estimation \cite{intro6}, and environmental pollution monitoring \cite{intro7}. Remote sensing imagery and label exhibits high heterogeneity across three key dimensions including spatial, temporal and spectral dimensions in dense prediction tasks, posing significant challenges for unified processing due to variations in image size, temporal length and spectral bands of input and output in practical applications \cite{intro8}.
    
    Remote sensing dense prediction refers to tasks where the goal is to produce a pixel-wise prediction, assigning a specific label to every pixel in the input satellite imagery \cite{dense_prediction_1}. Dense prediction in remote sensing primarily involves three core categories: semantic segmentation, binary change detection, and semantic change detection. These tasks can be formally defined as follows: given a remote sensing image time series of shape $(T_1, C_1, H_1, W_1)$, a dense prediction model is expected to produce a prediction of shape $(T_2, C_2, H_2, W_2)$, where $T_1$ and $T_2$ denote the temporal lengths of the input and output, $C_1$ is the number of input channels, $C_2$ is the number of output classes, and $(H_1, W_1)$ and $(H_2, W_2)$ denote the image sizes of the input and output, respectively. Typically, $H_1 = H_2$ and $W_1 = W_2$. From the task perspective, in semantic segmentation, the model aims to classify land cover types at a per-pixel level \cite{dense_prediction_1}, corresponding to $T_2=1$ and $C_2 \geq 2$; in binary change detection, the model identifies whether changes occur between adjacent time points \cite{fcef}, corresponding to $T_2 = T_1 - 1$ and $C_2 = 2$; in semantic change detection, the model extracts land cover information at each time step to analyze semantic differences between adjacent two time points \cite{scd}, corresponding to $T_2 = T_1$ and $C_2 \geq 2$. Therefore, the temporal dimension $T_1,T_2$ is correlated with the type of dense prediction tasks, and the output spectral dimension $C_2$ is correlated with the set of predicted semantic categories. From the data perspective, the input's spatial dimension is related to the geographical coverage and spatial resolution of remote sensing data; the input's temporal dimension correlates with the temporal coverage and temporal resolution of remote sensing data; the input's spectral dimension corresponds to the modality of remote sensing data. This underscores the importance of unified modeling of inputs and outputs in the spatial-temporal-spectral dimensions to achieve a unified approach for remote sensing dense prediction tasks.

    \begin{figure*}[!ht]
     \centering
     	\includegraphics[width=\linewidth]{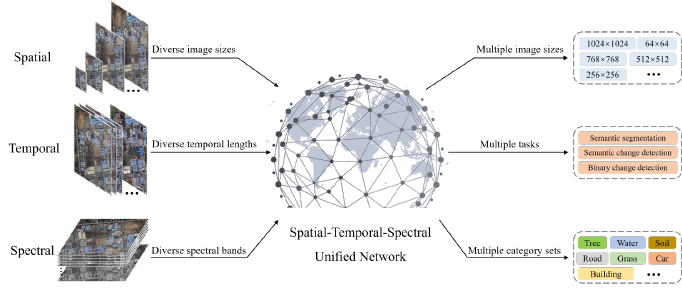}
     \caption{Illustration of Spatial-Temporal-Spectral Unified Network. STSUN is capable of handling input and output with arbitrary spatial, temporal and spectral dimension configurations, unifies semantic segmentation, binary change detection and semantic change detection tasks with flexible category set.}
     \label{Fig:intro}
    \end{figure*}

    While deep learning methods have achieved significant progress in remote sensing dense prediction, yielding high-performing models across semantic segmentation, binary change detection, and semantic change detection tasks \cite{sgsln, fccdn, rsmamba, bit}, current architectures exhibit several limitations. \textbf{1. Fixed Configurations}: These models are typically designed for fixed input-output configurations, defined by specific image sizes $((H_1,W_1),(H_2,W_2))$, temporal lengths $(T_1,T_2)$ and spectral bands $(C_1,C_2)$. This rigidity restricts their adaptability to the diverse and heterogeneous remote sensing datasets encountered in practical applications, such as diverse satellite sensor data with varying spectral bands or multi-temporal imagery sequences with varying temporal lengths. \textbf{2. Fixed Task:} These models neglect the inherent correlations among semantic segmentation, binary change detection, and semantic change detection tasks. This oversight often mandates the development of distinct models or dedicated task-specific decoders for each task, hindering unified processing pipelines. \textbf{3. Fixed Category Set:} Existing approaches are generally constrained to dense prediction with a fixed set of output categories. Consequently, any alteration to the target dense prediction schema, even a slight change, often result in substantial performance degradation or complete incompatibility, necessitating extensive retraining or fine-tuning, which incurs considerable computational and temporal overhead.
    
    To overcome these limitations, this study introduces the Spatial-Temporal-Spectral Unified Network (STSUN), as illustrated in Figure \ref{Fig:intro}. STSUN offers a novel framework for unified representation and modeling of remote sensing data across diverse spatial, temporal and spectral dimensions, solving the above issues in the following ways. \textbf{1. Unified Representation:} STSUN effectively addresses the variability in data characteristics by leveraging inherent metadata from each dimension to achieve a unified representation. This capability allows STSUN to seamlessly adapt to input and output data with arbitrary image sizes, temporal lengths, and spectral bands, thereby leveraging a wide range of data to build high-performance unified models for dense prediction. \textbf{2. Unified Tasks:} STSUN unifies semantic segmentation, binary change detection, and semantic change detection tasks within a single framework, which is accomplished by using a predefined and trainable task embedding set. By incorporating a selected task embedding and output temporal length into the output temporal dimension as metadata, the model is explicitly instructed to execute on the exact dense prediction task and the necessary output temporal length. This approach eliminates the need for separate models or task-specific decoders, enabling effective multi-task joint modeling and leveraging extensive data from multiple dense prediction tasks to improve model performance. \textbf{3. Flexible Category Set:} STSUN facilitates flexible prediction across multiple sets of semantic categories, which is accomplished by using a predefined and trainable category embedding set. The selected category subset is integrated into the output channel dimension as metadata, which guides the model to identify the specific classes needed for a given prediction. This enables the model to adapt to multiple subsets of prediction categories without fine-tuning, applying to diverse remote sensing scenarios with various semantic categories.
    
    To further improve the model's ability to capture diverse STS combinations, we design a Local-Global Window Attention (LGWA) mechanism. This module efficiently extracts local features using three overlapping window-based attention blocks with different shapes, followed by a global attention block that aggregates information at the global level. This design achieves a balance between computational efficiency and expressive power, enabling collaborative modeling of local and global features and boosting the model's performance in complex remote sensing tasks.

    The main contributions of this work are as follows:
    \begin{enumerate}
    \item We propose the Spatial-Temporal-Spectral Unified Network, which accommodates arbitrary input and output configurations across spatial, temporal, and spectral dimensions for the first time. This addresses the prevalent issue of model rigidity in handling heterogeneous remote sensing data.
    
    \item We introduce a unified task execution method that leverages trainable task embeddings to perform semantic segmentation, binary change detection, and semantic change detection within a single architecture, thereby obviating the need for separate models or task-specific heads.
    
    \item We design a flexible category prediction method, which utilizes trainable semantic embeddings to enable the model to perform dense prediction for multiple specified set of output classes, removing the constraint of a fixed category schema and the associated costs of retraining.
    
    \item We develop a Local-Global Window Attention mechanism that efficiently captures both local and global contextual features. This design enhances the model's feature extraction capabilities and improves performance across a wide range of remote sensing prediction tasks.
    \end{enumerate}

%
%
\section{Related Works}

\subsection{Data Dimensionality in Remote Sensing Dense Prediction}

    Remote sensing datasets utilized for dense prediction tasks are characterized by extensive variability across spatial, temporal and spectral dimensions. This heterogeneity stems from the diverse array of satellite and airborne sensors, each with unique acquisition parameters, mission objectives, and coverage patterns, tailored for different application scenarios and geographical regions \cite{related_24}. Such variability presents a formidable challenge for developing universally applicable and robust dense prediction models.

    Remote sensing imagery exhibits substantial spatial heterogeneity stemming from two key properties: ground sampling distance (GSD) and geospatial extent. The GSD is sensor-dependent, ranging from sub-meter resolutions ideal for detailed urban analysis \cite{pami_building_extract} to coarser resolutions suited for regional land cover mapping \cite{related_39}. The geospatial extent, meanwhile, is determined by the specific application, leading to variations in the captured area. Together, these properties dictate not only the pixel dimensions of an image but also the representation of its content. This variability poses a critical generalization challenge for deep learning models. A model trained for a specific resolution and image size typically suffers a significant performance degradation when applied to data with different spatial characteristics. This failure arises from two coupled effects: variations in GSD alter the perceived scale and texture of ground objects, while changes in geospatial extent modify the available contextual information.
    
    The temporal dimension in remote sensing data exhibits significant disparities. Revisit frequency, a critical factor for monitoring dynamic phenomena, ranges from multiple observations per day with sensors like MODIS \cite{related_25}, to several days (e.g., Sentinel-1 and Sentinel-2, with 5-12 day repeat cycles depending on latitude and constellation status \cite{related_27, related_28}), to 16 days for Landsat missions \cite{related_29}. Consequently, the temporal length of image sequences available for analysis can vary from bi-temporal pairs, commonly used in building change detection \cite{pami_cd, pami_non_dl_cd}, to dense time series comprising hundreds of observations, which are invaluable for applications like agricultural monitoring \cite{related_30}, vegetation forecasting \cite{vegediff} and moving object detection \cite{pami_rs_video}. Models trained on data with a specific temporal sequence length may not apply to datasets with different temporal characteristics without substantial retraining and adaptation.
    
    Spectral dimensionality is another source of major variation. The number of spectral bands can range from a single panchromatic band to a few multispectral bands (e.g., 4-bands in NAIP imagery, 13 bands in Sentinel-2 MSI \cite{related_28}), to hundreds of narrow, contiguous bands in hyperspectral sensors like AVIRIS \cite{related_32} or the upcoming EnMAP mission \cite{related_33}. Each sensor captures information from different portions of the electromagnetic spectrum, with varying band central wavelengths and bandwidths. This spectral diversity allows for the discrimination of different materials and land cover types based on their unique spectral signatures \cite{pami_spectralgpt}. However, it also means that models developed for one sensor may not be directly applicable to data from another sensor  without strategies to handle the differing spectral bandss and spectral information content \cite{related_34}. A series of models have been proposed to use hypernetworks to unify the inputs with different spectral bands, but ignored the unification of the output spectral bands \cite{dofa, wla}.
    
    The inherent heterogeneity in image size, temporal frequency, spectral composition across remote sensing datasets poses a substantial hurdle for developing universally applicable dense prediction models. Consequently, there is a pressing research gap concerning the development of adaptive model architectures or unified data processing strategies that can effectively ingest and interpret such diverse time-spectrum-space data formats at the input and output level for comprehensive and robust dense prediction.

\subsection{Task Unification in Remote Sensing Dense Prediction}

Dense prediction in remote sensing encompasses a range of pixel-level interpretation tasks that are crucial for remote sensing applications such as environmental monitoring, urban planning, and disaster assessment \cite{dense_prediction_1}. Among these, three tasks are fundamental: semantic segmentation, binary change detection, and semantic change detection. Semantic segmentation aims to assign a specific class label to every pixel in the satellite images, producing a detailed land cover map \cite{pami_seg}. Binary change detection, conversely, utilizes multi-temporal images to identify pixels where any form of change has occurred, outputting a binary map of {change, no change} \cite{fcef}. Bridging these two is semantic change detection, which not only detects changes across multi-temporal  images but also identifies the "from-to" nature of the change, such as a 'forest' pixel becoming a 'building' pixel \cite{sfccd}.

Given the significant conceptual overlap and inherent correlations among these tasks, a growing trend in the community is task unification through multi-task learning frameworks. Developing unified models to handle multiple dense prediction tasks simultaneously can lead to improved performance and efficiency \cite{sfccd, fccdn}, which leverage shared representations to allow complementary information from one task to benefit others. A model trained jointly for multiple tasks can learn more robust feature extractors than a model trained on either task alone \cite{pami_exchange_mtl}. For instance, SFCCD features task-specific branches for building semantic segmentation and change detection \cite{sfccd}. It leverages paired data from both tasks for training, resulting in superior change detection performance compared to training the change detection branch solely with change detection data. Similarly, FCCDN incorporates multi-task branches for semantic segmentation and change detection, and it utilizes the weakly supervised results from the semantic segmentation branch to enhance change detection effectiveness \cite{fccdn}. The core benefit of task unification is the potential to create more powerful and generalizable models while reducing the need to develop and deploy multiple specialized networks.

Despite their superior performance, existing multi-task unification efforts suffer from critical limitations that hinder their scalability and effectiveness. The first issue is a pervasive dependence on paired data. Current models typically require that the training datasets are fully annotated for all constituent tasks, which means the exact same set of images must possess corresponding pixel-level labels for semantic segmentation, binary change, and semantic change. This stringent requirement drastically limits the pool of usable training data, as such comprehensively annotated, multi-task datasets are exceedingly rare and expensive to create. The second issue is the common architectural choice of using task-specific decoders. Many unified models employ a shared encoder to extract features but diverge into separate, specialized decoder heads for each task. This design can create information bottlenecks and prevent the model from fully exploiting the synergies and complementarities among the tasks at the deepest levels of feature decoding and synthesis.

Consequently, there is an absence of a task unified dense prediction model that can effectively unify multiple core tasks through a joint modeling paradigm that operates on non-paired data.

\subsection{Flexible Class Sets in Remote Sensing Dense Prediction}

The objective of most remote sensing dense prediction tasks is to categorize ground objects according to a single predefined semantic class set. However, the definition of this class set is not universal, which is highly dependent on the specific application and the characteristics of the geographic scene. This variability is evident across different tasks and scenarios. For example, a simple building extraction task may only require a binary class set of {building, background}, while binary change detection operates on {change, no change}. In contrast, a standard Land Use/Land Cover (LULC) classification task might involve a more complex set, such as {water, forest, cropland, urban, barren}.

This variability is further compounded by scene-driven factors. The LULC class schema for a dense urban environment might need to include specific categories like {commercial building, residential building, road, playground, vehicle}, which would be irrelevant in a forest monitoring application. The latter might instead require a fine-grained class set like {coniferous forest, bamboo forest, tea plantation, shrubland}. While existing deep learning models have achieved impressive performance, they are almost universally designed for a specific and fixed class schema \cite{bit, sgsln, rsmamba}.

The primary limitation of these models is their inherent inflexibility with respect to the semantic class set. They are designed, trained, and optimized for a fixed vocabulary of predefined classes. The architecture of the model, particularly the final classification layer, is hard-wired to the number and identity of these classes. As a result, even a minor modification to the class set would render the pretrained model unusable for the new task, such as adding a 'wetland' category to a LULC model or splitting the 'building' class into 'residential' and 'commercial'. Adapting the model requires retraining or fine-tuning on a new dataset that includes the modified class set.

Therefore, there lacks a unified remote sensing dense prediction model capable of flexibly adapting to diverse class sets across different tasks and scenes without necessitating costly and time-consuming retraining.

%
%
\section{Methodology}

\subsection{Problem Formulation}

Dense prediction in remote sensing encompasses a range of tasks that, despite their distinct objectives, share a common foundation: inferring structured, pixel-wise semantic information from multi-dimensional earth observation data. We formalize these tasks within a unified mathematical framework, conceptualizing them as a highly flexible tensor-to-tensor mapping problem. This abstraction is essential for accommodating the inherent heterogeneity of remote sensing data in the spatial, temporal, and spectral dimensions.

Let an input remote sensing data instance be represented by a primary data tensor \( X \in \mathbb{R}^{T_1 \times C_1 \times H_1 \times W_1} \), while the corresponding model output is a prediction tensor \( Y \in \mathbb{R}^{T_2 \times C_2 \times H_2 \times W_2} \), which is the same as the formulation in Section \ref{sec:intro}. \( Y \) are typically aligned with the \( X \) by formulation (\(H_2 = H_1, W_2 = W_1\)) and specific dense prediction task, where the spatial, temporal, and spectral dimensions of \( X \) and \( Y \) can vary significantly, while \( T_1 \) and \( T_2 \) is strongly correlated with the dense prediction tasks, and \( C_2 \) is strongly correlated with the predicted set of semantic categories.

To address the profound variability in data characteristics, task and category requirements, we augment this core tensor representation with explicit metadata. The model's behavior is conditioned not only on the data tensor \( X \) but also on two metadata sets: an input set \( M_{in} \) describing the source data, and an output set \( M_{out} \) specifying the target prediction structure. These sets are decomposed as follows:
\begin{itemize}
    \item \textbf{Input Dimension Metadata:} \( M_{in} = \{M_{spa}^{in}, M_{tem}^{in}, M_{spe}^{in}\} \). This set provides essential context about the input data \( X \). \( M_{spa}^{in} \) encodes spatial information of input data, including pixel locations and spatial resolution. \( M_{tem}^{in} \) contains the acquisition timestamps for each of the \( T_1 \) temporal slices. \( M_{spe}^{in} \) specifies the wavelength for each of the \( C_1 \) spectral channels.

    \item \textbf{Output Dimension Metadata:} \( M_{out} = \{M_{spa}^{out}, M_{tem}^{out}, M_{spe}^{out}\} \). This set instructs the model on the desired output format and task. \( M_{spa}^{out} \) defines the pixel locations and spatial resolution of the prediction \( Y \). \( M_{tem}^{out} \) determines the temporal nature of the task. It includes the target temporal length \( T_2 \) and a trainable task embedding that specifies which dense prediction task to perform. \( M_{spe}^{out} \) defines the prediction's semantic space. It consists of a dynamically selectable subset of trainable embeddings corresponding to the \( C_2 \) target classes.

    \end{itemize}
    
    Within this framework, our goal is to learn a single, unified mapping function \( f_{\theta} \) parameterized by \( \theta \), which can adapt to arbitrary input and output configurations. The general mapping is defined as:
    \begin{equation}
       Y = f_{\theta}(X, M_{in}, M_{out}).
    \end{equation}
    The three core tasks of semantic segmentation, binary change detection, and semantic change detection and various semantic category sets are thus handled as specific instances of this general function. This metadata-driven formulation provides the mathematical basis for a model that is not constrained to fixed data structures, predefined tasks or predefined category sets, enabling a truly unified approach to dense prediction in remote sensing.

\subsection{Overview of the Spatial-Temporal-Spectral Unified Network} 

    \begin{figure*}[!ht]
     \centering
        \includegraphics[width=0.8\linewidth]{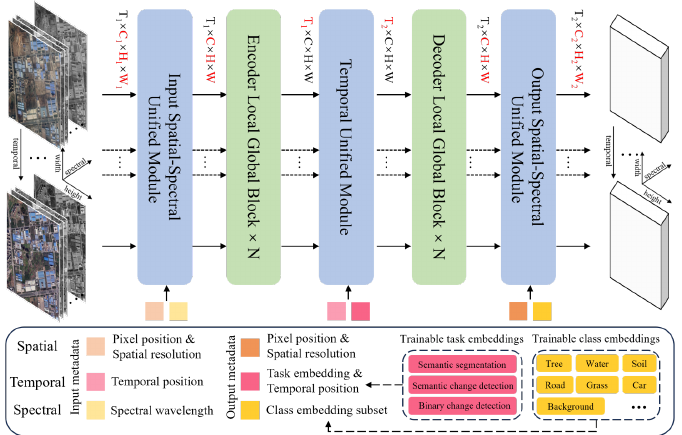}
        \caption{Overview of the STSUN architecture, comprising the Input Spatial-Spectral Unified Module for spatial and spectral unification of input, the Local Global Basic blocks for global and local feature extraction, the Temporal Unified Module for temporal unification of input and output,  and the Output Spatial-Spectral Unified Module for spatial and spectral unification of output. The highlighted portions in the shapes on either side of each unification module indicate the dimension that the module unifies.}
     \label{Fig:stsun}
    \end{figure*}

The Spatial-Temporal-Spectral Unified Network unifies data representation by leveraging metadata from input and output data across spatial, temporal, and spectral dimensions. It employs trainable task embeddings and class embeddings to specify the particular dense prediction task and the semantic classes set. This design enables STSUN to adapt to various input and output shape configurations and unifies semantic segmentation, binary change detection, and semantic change detection tasks, supporting predictions across diverse class sets. STSUN primarily consists of five stages, as illustrated in Figure \ref{Fig:stsun}.

First, at the input stage, the input spatial-spectral unified module (ISSUM) leverages the input spatial metadata \(M_{spa}^{in}\) and spectral metadata \(M_{spe}^{in}\) for unified encoding of spatial and spectral dimensions. The temporal dimension, being intrinsically linked to the specific dense prediction task, necessitates a distinct processing approach from the spatial and spectral dimensions. If the temporal dimension were to be unified at this stage, it would be equivalent to a pixel-level fusion of multi-temporal remote sensing images. This would introduce significant interference from irrelevant information, thereby degrading the model's performance on dense prediction tasks \cite{fccdn, sgsln}. Therefore, at this stage, the independence of the temporal dimension is preserved while the spatial and spectral dimensions are unified. In the spatial dimension, the spatial resolutions of the input images and the spatial position of each pixel are utilized as metadata for the input spatial dimension. This enables a uniform representation of data based on the similarity of spatial proximity, accommodating varied resolutions. Concurrently, in the channel dimension, the spectral wavelength of each spectral band is incorporated as metadata for the input channel dimension, facilitating a unified data representation grounded in spectral continuity.

Second, at the encoder stage, the Encoder Local Global Blocks are used to extract features from input data. As the temporal independence of remote sensing images is explicitly preserved, a shared-weight encoder then processes each temporal remote sensing image independently, extracting local and global features across distinct temporal instances within Encoder Local Global Blocks. The overall structure of the Encoder Local Global Blocks is the same as the transformer block, except that the global attention is replaced with local-global window attention, as shown in Figure \ref{Fig:LGWA}.

Third, at the encoder–decoder junction, the temporal unified module (TUM) utilizes the input temporal metadata \(M_{tem}^{in}\) to fuse features from different temporal instances and adjusts these features using the output temporal metadata \(M_{tem}^{out}\) to adapt them for the specific dense prediction task. The temporal position within the remote sensing image time series is used as metadata for the input time dimension, which enables feature-level fusion of remote sensing image features across various time points, effectively mitigating interference from irrelevant temporal information. Simultaneously, to guide the dense prediction task, multiple predefined trainable task embeddings are introduced. A specific task embedding is selected and incorporated as metadata with output temporal length for the output time dimension, thereby instructing the model on the required dense prediction task. It is crucial to note that operations performed in this part maintain temporal independence and do not merge with the channel dimension \cite{fccdn, sgsln}. This approach allows different dense prediction tasks to be unified within an independent temporal dimension, preventing interference from the remote sensing image features. Instead, it guides the subsequent decoder to optimize these features for the specified task, which is conducive to the model's ability to jointly model multiple dense prediction tasks.

Fourth, the decoder subsequently processes the fused multi-temporal remote sensing image features. Since the features from multiple temporal instances have been fused and the model has been instructed on the dense prediction task via the task embedding, the shared-weight decoder local-global blocks optimize the features for the specific task by removing irrelevant information, progressively aligning them with the target output. The Decoder Local Global Block are the same as Encoder Local Global Block.

Fifth, in the output stage, the output spatial-spectral unified module (OSSUM) utilizes the output spatial metadata \(M_{spa}^{out}\) to restore the features to the original image dimensions and leverages the output spectral metadata \(M_{spe}^{out}\) to guide the features toward making dense predictions for a specific subset of semantic categories. In the spatial dimension, the spatial resolution of the remote sensing images and the patch position information are again leveraged as metadata for the output spatial dimension to accurately restore features to the original input image size. In the channel dimension, multiple predefined trainable category embeddings are established. From these, a subset of category embeddings is dynamically selected and integrated as metadata for the output channel dimension, explicitly indicating the set of categories the model is instructed to predict, thus enabling flexible class-agnostic prediction.

Through the arrangement of spatial, temporal, and spectral dimensions across these five stages, and the design of metadata for each dimension, STSUN accommodates diverse input/output dimensional configurations and various subsets of prediction classes. Moreover, these designs enable the effective joint modeling of semantic segmentation, binary change detection, and semantic change detection. By uncovering the correlations and complementarities among these tasks, STSUN leverages the combined data of multiple tasks to enhance its performance across multiple dense prediction challenges.

\subsection{Spatial-Temporal-Spectral Unified Module}  

    \begin{figure*}[!ht]
     \centering
        \includegraphics[width=\linewidth]{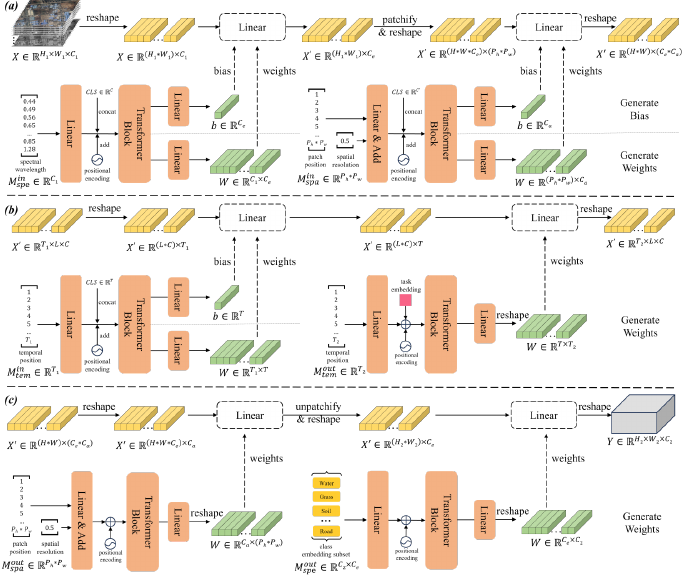}
        \caption{Illustration of the Spatial-Temporal-Spectral Unified Module. (a) Input Spatial-Spectral Unified Module. (b) Temporal Unified Module. (c) Output Spatial-Spectral Unified Module.}
     \label{Fig:STSUM}
    \end{figure*}

To facilitate the mapping from the raw data space to a unified feature space across different dimensions (spatial, temporal, and spectral), and to transform these unified features into outputs of appropriate shapes according to specific task and class requirements, our proposed STSUN employs the Spatial-Temporal-Spectral Unified Module (STSUM) to handle STS dimension, which consists of ISSUM, TUM and OSSUM . First, the ISSUM maps the variable input image size $(H_1, W_1)$ and spectral bands $C_1$ to a predefined, unified size $(H, W)$ and spectral bands $C$ in the spatial and spectral dimensions, respectively. Subsequently, the TUM maps the variable input temporal length $T_1$ to a predefined, unified length $T$, which is then mapped to a variable output temporal length $T_2$ based on task demands. Finally, the OSSUM maps the unified image size $(H, W)$ and spectral bands $C$ to a variable output size $(H_2, W_2)$ and spectral bands $C_2$, guided by the prediction class set and image size requirements. The details of these modules are illustrated in Figure \ref{Fig:STSUM} (a), (b), and (c).

The core mechanism of STSUM involves using metadata from each dimension, along with optional trainable embeddings, to generate adaptive linear layers. This enables the transformation of variable input data into unified features or, conversely, the conversion of unified features into variable outputs tailored to specific requirements. The feature mapping mechanism is consistent across the modules and comprises two main branches: a hyper-network branch and a mapping network branch. In the hyper-network branch, metadata from each dimension is first tokenized using a linear layer and augmented with positional encodings. These tokens are then processed through several Transformer blocks to capture the latent relationships among them. Finally, another linear layer generates the parameters for the adaptive mapping network. In the mapping network branch, this dynamically generated network applies a linear transformation to the input features, thereby unifying the input data or generating the adaptive output. Although the underlying principle is similar, each module requires distinct and meticulous design considerations regarding feature shapes and parameter generation. This ensures that the modules can achieve generic and robust dimensional unification by processing different metadata for different mapping requirements across various dimensions.

\subsubsection{Input Spatial-Spectral Unified Module}

The ISSUM unifies the spectral and spatial dimensions of the input data. It first operates on the spectral dimension, mapping the variable input spectral bands $C_1$ to a predefined, unified spectral bands $C$ using spectral metadata. It then proceeds to the spatial dimension, mapping the variable input image size $(H_1, W_1)$ to a predefined, unified size $(H, W)$ using spatial metadata, as depicted in Figure \ref{Fig:STSUM}(a). For a batch of input data $X \in \mathbb{R}^{T_1 \times H_1 \times W_1 \times C_1}$, since the ISSUM is designed to unify the spatial and spectral dimensions, the temporal dimension $T_1$ can be fused with the batch dimension. Consequently, the ISSUM only needs to process a single time-step input $X \in \mathbb{R}^{H_1 \times W_1 \times C_1}$. For the spectral dimension of $X$, the hyper-network branch utilizes the input data's spectral wavelengths as metadata $M_{\text{spe}}^{\text{in}} \in \mathbb{R}^{C_1}$. This metadata is first tokenized to $M' \in \mathbb{R}^{C_1 \times C_e}$ by a linear layer, where $C_e$ is a predefined unified spectral bands. Subsequently, a learnable class token, $\text{CLS} \in \mathbb{R}^{C_e}$, is concatenated to the token sequence, and positional encodings are added to incorporate relative position information, resulting in $M' \in \mathbb{R}^{(C_1+1) \times C_e}$. $M'$ is then processed by multiple Transformer blocks to capture latent relationships among the different spectral bands. The output is split into two parts: the $\text{CLS}$ token and the remaining token sequence $M''$. The $\text{CLS}$ token is passed through a linear layer to generate the bias parameter $b \in \mathbb{R}^{C_e}$, while $M''$ is passed through another linear layer to generate the weight matrix $W \in \mathbb{R}^{C_1 \times C_e}$. The generated $W$ and $b$ constitute a linear layer capable of mapping input features with $C_1$ channels to output features with $C_e$ channels. Accordingly, in the mapping network branch, the ISSUM reshapes $X \in \mathbb{R}^{H_1 \times W_1 \times C_1}$ to $X \in \mathbb{R}^{(H_1 \cdot W_1) \times C_1}$ and then transforms it using the generated mapping network to obtain $X' \in \mathbb{R}^{(H_1 \cdot W_1) \times C_e}$, thus achieving unification in the channel dimension.

Next, for the spatial dimension of $X'$, the ISSUM begins in the mapping network branch by applying `patchify` and `reshape` operations to $X' \in \mathbb{R}^{(H_1 \cdot W_1) \times C_e}$, transforming it into $X' \in \mathbb{R}^{(H \cdot W \cdot C_e) \times (P_h \cdot P_w)}$, where $P_h = H_1/H$ and $P_w = W_1/W$. This procedure converts the variable image size $(H_1, W_1)$ into a unified spatial size $(H, W)$ by moving the variable part into the channel dimension, which allows the ISSUM to unify a variable spectral bands of $P_h \cdot P_w$ in a similar way. Consequently, in the hyper-network branch, the ISSUM uses the patch positions and spatial resolution of the input data as spatial metadata, $M_{\text{spa}}^{\text{in}} \in \mathbb{R}^{P_h \cdot P_w}$. These are tokenized by separate linear layers and then summed to form the metadata token sequence, where the patch position denotes the location information of each pixel within a single patch. Finally, employing the same procedure as described for the spectral dimension, the ISSUM uses this spatial metadata token sequence to generate a bias parameter $b \in \mathbb{R}^{C_a}$ and a weight matrix $W \in \mathbb{R}^{(P_h \cdot P_w) \times C_a}$, where the $C_a$ is predefined unified spectral bands. These parameters define a linear layer that maps $X' \in \mathbb{R}^{(H \cdot W \cdot C_e) \times (P_h \cdot P_w)}$ to $X' \in \mathbb{R}^{(H \cdot W \cdot C_e) \times C_a}$. After reshaping, this yields $X' \in \mathbb{R}^{(H \cdot W) \times (C_e \cdot C_a)}$, thereby completing the unification of the input data in the spatial dimension.

\subsubsection{Temporal Unified Module}

The TUM unifies the temporal dimension of the input and output data. It employs a similar mechanism with ISSUM to first map the variable input temporal length $T_1$ to a predefined, unified length $T$, and then map this unified length to a variable output length $T_2$ based on task requirements, as shown in Figure \ref{Fig:STSUM}(b). For simplicity, let $C = C_e \cdot C_a$, $L = H \cdot W$. The TUM isolates the temporal dimension, obtaining a single data sample $X' \in \mathbb{R}^{T_1 \times L \times C}$, and reshapes it to $X' \in \mathbb{R}^{(L \cdot C) \times T_1}$, thereby transposing the temporal dimension to the position of the channel dimension. This allows the TUM to unify the temporal dimension of the input data using a similar method of ISSUM. TUM utilizes the temporal position information of the input data as metadata, $M_{\text{tem}}^{\text{in}} \in \mathbb{R}^{T_1}$, to adaptively generate a linear layer composed of a bias parameter $b \in \mathbb{R}^{T}$ and a weight matrix $W \in \mathbb{R}^{T_1 \times T}$. This mapping network transforms the temporally variable input $X' \in \mathbb{R}^{(L \cdot C) \times T_1}$ into a unified representation $X' \in \mathbb{R}^{(L \cdot C) \times T}$. 

Subsequently, according to the requirements of the specific dense prediction task, the TUM needs to map the unified temporal length $T$ to a variable length $T_2$. Unlike the previous operation of mapping variable features to unified features, this process is reversed, which necessitates differences in the metadata and the generated parameters. Specifically, the TUM uses the output temporal length information and a selected task embedding as the output temporal metadata, $M_{\text{tem}}^{\text{out}} \in \mathbb{R}^{T_2}$. These are mapped through linear layers and then summed to form the metadata token sequence. The task embedding is a predefined, trainable embedding selected from a set, such as \{semantic segmentation embedding, binary change detection embedding, semantic change detection embedding\}, to specify the dense prediction task being performed. The metadata token sequence is then processed by a Transformer block to capture latent relationships between tokens. Following this, it passes through a linear layer to generate only the weight parameter $W \in \mathbb{R}^{T_2 \times T}$, which is reshaped to $W \in \mathbb{R}^{T \times T_2}$ to serve as the weights of the mapping network. Finally, this bias-free mapping network transforms $X' \in \mathbb{R}^{(L \cdot C) \times T}$ into $X' \in \mathbb{R}^{(L \cdot C) \times T_2}$, which is then reshaped back to $X' \in \mathbb{R}^{T_2 \times L \times C}$, thus converting the unified temporal length $T$ into a variable length $T_2$ according to specific task demands. The bias parameter is not generated because the mapping network would require a variable bias $b \in \mathbb{R}^{T_2}$, which cannot be generated from a fixed $\text{CLS}$ token through a fixed linear layer.

\subsubsection{Output Spatial-Spectral Unified Module}

The OSSUM unifies the spatial and spectral dimensions of the output data. It first maps the unified image size $(H, W)$ to a variable size $(H_2, W_2)$ using spatial metadata, and then maps the unified spectral spectral bands $C$ to a variable count $C_2$ using spectral metadata, as illustrated in Figure \ref{Fig:STSUM}(c). The dimensional mapping mechanism of the OSSUM is similar to that of the TUM for the output temporal dimension, with the main difference lying in the metadata. Specifically, for the spatial dimension, the hyper-network branch of the OSSUM uses the patch position and spatial resolution as metadata, $M_{\text{spa}}^{\text{out}} \in \mathbb{R}^{P_h \cdot P_w}$. After passing through linear layers and Transformer blocks, it generates a weight parameter $W \in \mathbb{R}^{C_a \times (P_h \cdot P_w)}$ for the mapping network. Since the output spatial size is identical to the input spatial size, we have $M_{\text{spa}}^{\text{out}} = M_{\text{spa}}^{\text{in}}$. In the mapping network branch, the input feature $X' \in \mathbb{R}^{(H \cdot W) \times (C_e \cdot C_a)}$ is reshaped to $X' \in \mathbb{R}^{(H \cdot W \cdot C_e) \times C_a}$. It is then transformed by the mapping network to yield $X' \in \mathbb{R}^{(H \cdot W \cdot C_e) \times (P_h \cdot P_w)}$. An `unpatchify` operation performs up-sampling, and a final reshape operation converts it to $X' \in \mathbb{R}^{(H_2 \cdot W_2) \times C_e}$, producing a variable-sized output in the spatial dimension.

Similarly, in the spectral dimension, the OSSUM first selects a subset from a predefined set of semantic class embeddings based on the requirements of the prediction class set. This subset of embeddings indicates all the classes the model needs to predict. For instance, while multiple remote sensing scenes might require dense prediction for various land cover types, leading to a total semantic class embedding set like \{Tree, Water, Soil, Road, Building, Background\}, a specific building extraction task would only require selecting the subset \{Building, Background\}. This subset directs the model to classify land cover into these two categories, enabling effective building extraction. Therefore, the selected subset of semantic class embeddings serves as the output spectral metadata, $M_{\text{spe}}^{\text{out}} \in \mathbb{R}^{C_2 \times C_e}$. Through a similar mechanism above, a weight parameter $W \in \mathbb{R}^{C_e \times C_2}$ is generated for the mapping network. In the mapping network branch, $X' \in \mathbb{R}^{(H_2 \cdot W_2) \times C_e}$ undergoes a linear transformation by the mapping network to produce $X' \in \mathbb{R}^{(H_2 \cdot W_2) \times C_2}$. Finally, a reshape operation yields the output $Y \in \mathbb{R}^{H_2 \times W_2 \times C_2}$, producing a variable output in the spectral dimension.

\subsection{Local-Global Window Attention}  

\begin{figure}[!ht]
     \centering
        \includegraphics[width=\linewidth]{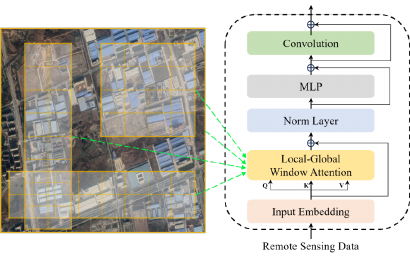}
        \caption{Architecture of the LGWA-based Local Global Block, employing multiple local windows to extract local feature and single global window to extract global feature.}
     \label{Fig:LGWA}
    \end{figure}

After unifying the diverse spatial-temporal-spectral input and output data, STSUN requires a powerful feature extraction capability to be effectively applied to various remote sensing scenarios and tasks. Therefore, this study proposes the Local-Global Window Attention module. This module performs self-attention operations within multiple local windows of varying shapes and a single global window, As illustrated in Figure~\ref{Fig:LGWA}. This design enables the joint capture of fine-grained local details and coarse-grained global features, thereby effectively extracting the local characteristics of various ground objects and modeling the global contextual relationships among them, which enhances the model's robustness across diverse remote sensing applications.

The LGWA module incorporates three local window attention mechanisms with varying sizes and configurations, alongside a single global attention mechanism. The distinct shapes of the local windows are specifically designed to extract different levels of local features. The horizontal window concentrates on capturing horizontally-oriented local details, the vertical window focuses on vertically-aligned local features, and the rectangular window is dedicated to extracting omnidirectional local information. This multi-faceted approach ensures the comprehensive extraction of multi-level features from ground objects. Subsequently, the global attention mechanism models the global contextual relationships among these objects based on their extracted local features, which enhances the overall performance of the model.

    For a specific attention window, the input sequence \( X \in \mathbb{R}^{L \times d_M} \) is projected into query, key, and value matrices \( Q, K, V \) through linear projections, formulated as:  
    \begin{equation}
        Q = XW^Q, \quad K = XW^K, \quad V = XW^V,
    \end{equation}  
    where \( W^Q, W^K \in \mathbb{R}^{d_M \times d_k} \) and \( W^V \in \mathbb{R}^{d_M \times d_v} \) are learnable weight matrices. The attention calculation within each window focuses on extracting fine-grained features, while different window configurations provide sensitivity to varying scales.

    The attention scores are computed using the scaled dot-product attention mechanism:  
    \begin{equation}
    \text{Attn}(Q, K, V) = \text{softmax}\left( \frac{QK^T}{\sqrt{d_k}} \right) V,
    \end{equation}  
    
    To further enhance the feature representation, the LGWA module employs the multi-head attention strategy, allowing each head to independently execute the above process and then concatenate the results:  
    \begin{equation}
    \text{MultiHead}(Q, K, V) = \text{Concat}(H_1, H_2, \ldots, H_h) W^O,
    \end{equation}  
    where \( H_i = \text{Attn}(QW_i^Q, KW_i^K, VW_i^V) \), and \( W^O \in \mathbb{R}^{hd_v \times d_M} \) projects the concatenated output back to the original dimension.

%
%

\section{Experiments and Results}

To validate the adaptability of STSUN to arbitrary Spatial-Temporal-Spectral inputs and outputs, and its capability to concurrently perform semantic segmentation, binary change detection, and semantic change detection tasks with support for variable class subsets, we conducted experiments on a total of six datasets across building and Land Use/Land Cover (LULC) scenarios. For each scenario, a unified STSUN model was trained using the combined training sets from all datasets within that scenario, and its performance was evaluated on their respective test sets.

In the building scenario, we selected the WHU, WHU-CD, LEVIR-CD, and TSCD datasets. While these datasets share the same number of channels in their input and label data, they exhibit variations in temporal and spatial dimensions, leading to differences in dense prediction task types, thus verifying that STSUN can adapt to various spatiotemporal input and output settings, and can unify semantic segmentation, binary change detection, and semantic change detection  tasks. Specifically, the WHU dataset corresponds to a single-temporal building semantic segmentation task, the WHU-CD dataset is associated with a bi-temporal building semantic change detection task, the LEVIR-CD dataset is used for a bi-temporal building binary change detection task, and the TSCD dataset pertains to a multi-temporal building binary change detection task, as summarized in Table \ref{table: dataset}.

In the LULC scenario, we chose the LoveDA Urban and Dynamic EarthNet datasets. These datasets differ in their temporal and spectral dimensions for both input and label data, which verifies that STSUN can adapt to various spatial-temporal and spectral input and output settings, unify semantic segmentation, binary change detection, and semantic change detection  tasks, and support dense prediction tasks with varying semantic class sets. The LoveDA Urban dataset corresponds to a single-temporal semantic segmentation task with an semantic category subset of 7 LULC classes, while the Dynamic EarthNet dataset is used for multi-temporal binary change detection and semantic change detection tasks with another semantic category subset of 6 LULC classes, as detailed in Table \ref{table: dataset}.


\begin{table*}[!ht]
\caption{Brief Introduction of The Experimental Datasets.}
\label{table: dataset}
\centering
\begin{tabular}{lccccccccc}
\hline
Name            & Scenario    & Task                      & $T_1$ & $T_2$ & $C_1$ & $C_2$ & Image Size  & Resolution & Images \\ \hline
WHU \cite{whu_dataset}            & Building & SS     & 1       & 1        & 3       & 1        & 512×512     & 0.3        & 8189   \\
WHU-CD \cite{whu_dataset}         & Building & SCD & 2       & 2        & 3       & 1        & 1024×1024 & 0.075      & 480      \\
LEVIR-CD \cite{levircd_dataset}       & Building & BCD   & 2       & 1        & 3       & 1        & 1024×1024   & 0.5        & 445    \\
TSCD \cite{tscd_dataset}           & Building & BCD   & 4       & 3        & 3       & 1        & 256×256     & 0.5        & 2700   \\
LoveDA Urban \cite{loveda}   & LULC     & SS     & 1       & 1        & 3       & 7        & 1024×1024   & 0.3        & 5987   \\
Dynamic Earthnet \cite{dynamicearthnet} & LULC     & BCD \&SCD & 24      & 24       & 4       & 6        & 1024×1024   & 3          & 54750  \\ \hline
\end{tabular}
\end{table*}

\subsection{Datasets}

We offer a brief description of the experimental building and LULC scenario datasets in Table \ref{table: dataset}.

1) Building scenario datasets: The WHU Building dataset \cite{whu_dataset} is divided into two main components: one containing satellite imagery and another composed of aerial photos. In our study, we utilize the aerial photo subset, which consists of 8,189 images. These images are split into 4,736 for training, 1,036 for validation, and 2,416 for testing, each with a spatial resolution of 0.3 meters. In total, this subset represents over 22,000 buildings covering an area in excess of 450 square kilometers. Our experiments were conducted using the original partitioning scheme and image dimensions (512×512) as specified by the WHU dataset.

The WHU-CD dataset \cite{whu_dataset} includes bitemporal very high-resolution (VHR) aerial images taken in 2012 and 2016, which clearly highlights major changes in building structures. The dataset is partitioned into non-overlapping patches of 1024×1024 pixels. These patches are further allocated into training, validation, and test sets following a 7:1:2 ratio.

The LEVIR-CD dataset \cite{levircd_dataset} is an extensive resource for change detection, comprising VHR Google Earth images with a resolution of 0.5 m/pixel. These images capture a variety of building transformations over periods ranging from 5 to 14 years, with a particular emphasis on construction and demolition events. The bitemporal images have been expertly annotated using binary masks, where a label of 1 denotes a change and 0 signifies no change. In total, there are 31,333 labeled instances of building modifications. Our experimental setup used the dataset's original image dimensions of 1024×1024 and adhered to the provided data partitioning scheme.

The TSCD dataset \cite{tscd_dataset} is constructed from WorldView-2 satellite imagery with a spatial resolution of approximately 0.5 m/pixel, acquired in 2016, 2018, 2020, and 2022. To mitigate external influences, the images underwent co-registration using manually selected control points and resampling to ensure a consistent coordinate framework. Building footprints were densely labeled for each temporal phase. Subsequently, three sets of change labels (2016–2018, 2018–2020, 2020–2022) were generated by performing differential operations on adjacent building distribution maps. The final TSCD dataset was created through uniform cropping and partitioning of these original images and derived labels.

2) LULC scenario datasets: The LoveDA dataset \cite{loveda} consists of 5,987 high-resolution optical remote sensing images (with a ground sampling distance of 0.3 m) each sized at 1024×1024 pixels. It covers seven land cover classes: building, road, water, barren, forest, agriculture, and background. The dataset is divided into 2,522 training images, 1,669 images for validation, and 1,796 images for testing, all drawn from two distinct scenes—urban and rural—from three Chinese cities: Nanjing, Changzhou, and Wuhan. The dataset poses considerable challenges due to the presence of multiscale objects, complex backgrounds, and uneven class distribution.

The DynamicEarthnet dataset \cite{dynamicearthnet} comprises 55 daily Sentinel-2 Image Time Series (SITS) collected globally between January 1, 2018, and December 31, 2019. For each month, data from the first day is annotated, which results in 24 ground truth segmentation maps per Area of Interest (AoI). Each image is 1024×1024 pixels and multi-spectral, containing four channels (RGB plus near-infrared). The annotations cover general land-use and land-cover categories: impervious surface, agriculture, forest, wetlands, soil, and water. The 'snow' class appears in only a few AoIs and has been excluded from this study.

\subsection{Baseline}

To evaluate the effectiveness of the proposed STSUN, we conducted comparative experiments with various benchmark methods on building and LULC scene datasets. Since STSUN is adaptable to multiple datasets, it was trained across all datasets in each scenario, whereas the benchmark methods were trained on individual datasets.


On the four building scene datasets, the compared CNN-based models include FCN \cite{fcn}, SegNet \cite{segnet}, U-Net \cite{unet}, PSPNet \cite{psp}, HRNet \cite{hrnet}, MA-FCN \cite{mafcn}, Deeplabv3+ \cite{deeplab}, ResUNet-a \cite{resunet_a}, MAPNet
\cite{mapnet}, D-LinkNet \cite{dlinknet}, SIINet \cite{siinet}, FC-EF \cite{fcef}, FC-Siam-Diff \cite{fcef}, FC-Siam-Conc \cite{fcef},
STANet \cite{levircd_dataset}, DTCDSCN \cite{dtcdstn}, SNUNet \cite{snu}, CDNet \cite{cdnet}, DDCNN \cite{ddcnn}, DASNet \cite{dasnet}, DSIFN \cite{dsifn}, HANet \cite{hanet}, USSFCNet \cite{ultralightweight}, and SEIFNet \cite{spatiotemporal}, the compared transformer-based models include Segformer \cite{segformer}, ChangeFormer \cite{changeformer} and A2Net \cite{lightweight}, and the CNN-transformer hybrid models include BDTNet \cite{bdtnet}, TransUNet \cite{transunet}, CMTFNet \cite{cmtfnet}, BIT \cite{bit}, MTCNet \cite{mtcnet}, MSCANet \cite{mscanet}, AMTNet-50 \cite{amtnet}, Contrast-COUD \cite{tscd_dataset} and TS-COUD \cite{tscd_dataset}.

On the two LULC scene datasets, the compared CNN-based models include U-Net \cite{unet}, Deepabv3+ \cite{deeplab}, DANet \cite{danet}, ResUNet-a \cite{resunet_a}, DASSN \cite{dassn}, HCANet \cite{hcanet}, RAANet \cite{raanet}, MSAFNet \cite{msafnet}, A2-FPN \cite{a2fpn} and CAC \cite{cac}, the compared transformer-based models include LANet \cite{lanet}, SCAttNet \cite{scattnet}, CLCFormer \cite{clcformer} and TSViT \cite{tsvit}, the CNN-transformer hybrid models include SAPNet \cite{sapnet}, SSCBNet \cite{sscbnet}, UTAE \cite{utae}, A2Net \cite{a2net}, SCanNet \cite{scannet} and TSSCD \cite{tsscd}.

\subsection{Implementation Details}

1) Data Augmentation: To validate the proposed methods, we adopted a minimalistic yet effective data augmentation strategy, deliberately refraining from complex augmentation schemes. Specifically, the employed transformations were limited to horizontal/vertical flipping (probability = 0.5) and transposition (probability = 0.5).

2) Training and Inference: The STSUN model was implemented using PyTorch \cite{pytorch} and executed on a single RTX A100 GPU (80G). Due to the heterogeneous image resolutions across the datasets, the batch size was set to 16 for the four building scene datasets and 4 for the two LULC scene datasets. Our optimization strategy combined binary cross-entropy loss with Dice coefficient loss, facilitating a balanced performance optimization. The AdamW optimizer \cite{adam} was initialized with a learning rate of 0.0001 and a weight decay of 0.001. A learning rate scheduler was employed to reduce the learning rate by a factor of 0.1 if no increase in the mean F1-score was observed on the aggregate validation set for 5 consecutive epochs. The training process spanned 100 epochs, ensuring robust convergence, and the best performing checkpoints—corresponding to the maximum mean F1-scores achieved—were retained for the testing phase. Furthermore, in order to ensure comparability with existing methodologies, all models were initialized using the default PyTorch settings across all datasets.

3) Evaluation Metrics: The performance of the proposed models was quantitatively assessed using five principal metrics: overall accuracy (OA), precision (P), recall (R), F1-score, and intersection over union (IoU). For multi-temporal tasks and multi-category tasks, the average F1-score (AF) and mean IoU (mIoU) will be used. In addition, following the settings of the DynamicEarthnet dataset \cite{dynamicearthnet}, we use the semantic change segmentation (SCS) metric, classagnostic binary change score (BC) and semantic segmentation score among changed pixels (SC) metrics to evaluate the model's performance on this dataset. 


\subsection{Overall Comparison on the Building Scenario}

The efficacy of the proposed Spatial-Temporal-Spectral Unified Network was evaluated through extensive experiments on four benchmark remote sensing datasets: the WHU dataset for single-temporal semantic segmentation, the WHU-CD dataset for bi-temporal semantic change detection, the LEVIR-CD dataset for bi-temporal binary change detection, and the TSCD dataset for multi-temporal binary change detection. STSUN was compared against several state-of-the-art approaches, with quantitative results summarized in Table \ref{whu_result}, \ref{whucd_result}, \ref{levircd_result} and \ref{tscd_result}.

\begin{figure*}[!ht]
 \centering
 	\includegraphics[width=\linewidth]{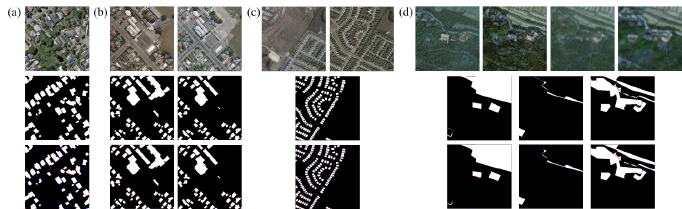}
 \caption{Sample inference results on for building scene datasets. The input images, ground truths and predictions are shown in the first, second and third rows, respectively. Red areas denote false positives and blue areas denote false negatives. (a) WHU dataset sample. (b) WHU-CD dataset sample. (c) LEIVR-CD dataset sample. (d) TSCD dataset sample.}
 \label{Fig:building_result}
\end{figure*}

On the WHU dataset, characterized by its complex building footprints and significant variations in object scale, our proposed STSUN achieves state-of-the-art performance. As detailed in Table \ref{whu_result}, STSUN obtains the highest IoU of 91.00\% and an F1-score of 95.29\%. This superior performance can be attributed to STSUN, particularly its spatial unification component, which effectively processes varying image size, and the Local-Global Window Attention mechanism. The LGWA module, with its ability to capture both fine-grained local details and broader contextual information, is particularly adept at delineating intricate building boundaries and accurately segmenting buildings of diverse sizes.

\begin{table}[!ht]
\caption{Accuracy Comparison on the WHU Dataset. The Best Values Are Highlighted in Bold.}
\label{whu_result}
\centering
\begin{tabular}{lcccc}
\toprule
Methods & P (\%) & R (\%) & F1 (\%) & IoU (\%) \\
\midrule
FCN~\cite{fcn} & 92.29 & 92.84 & 92.56 & 86.16 \\
SegNet~\cite{segnet} & 93.42 & 91.71 & 92.56 & 86.15 \\
U-Net~\cite{unet} & 94.50 & 90.88 & 92.65 & 86.31 \\
PSPNet~\cite{psp} & 93.19 & 94.21 & 93.70 & 88.14 \\
HRNet~\cite{hrnet} & 91.69 & 92.85 & 92.27 & 85.64 \\
MA-FCN~\cite{mafcn} & 94.75 & 94.92 & 94.83 & 90.18 \\
Deeplabv3+~\cite{deeplab} & 94.31 & 94.53 & 94.42 & 89.43 \\ 
ResUNet-a~\cite{resunet_a} & 94.49 & 94.71 & 94.60 & 89.75 \\
MAP-Net~\cite{mapnet} & 93.99 & 94.82 & 94.40 & 89.40 \\
Segformer~\cite{segformer} & 94.72 & 94.42 & 94.57 & 89.70 \\ 
TransUNet~\cite{transunet} & 94.05 & 93.07 & 93.56 & 87.89 \\
CMTFNet~\cite{cmtfnet} & 90.12 & \textbf{95.21} & 92.59 & 86.21 \\
\midrule
STSUN & \textbf{95.71} & 94.87 & \textbf{95.29} & \textbf{91.00} \\
\bottomrule
\end{tabular}
\end{table}

For the bi-temporal semantic change detection task on the WHU-CD dataset, STSUN demonstrates leading performance against other SOTA methods, as shown in Table \ref{whucd_result}. The WHU-CD dataset demands accurate identification of 'from-to' semantic transitions between two time points. STSUN excels here due to its inherent design for unified temporal and spectral  modeling. The STSUN component allows the network to effectively learn temporal evolutionary patterns and relationships between buildings and other objects, leading to more precise change maps.

\begin{table}[!ht]
\caption{Accuracy Comparison on the WHU-CD Dataset. The Best Values Are Highlighted in Bold.}
\label{whucd_result}
\centering
\begin{tabular}{lcccc}
\toprule
Methods & P (\%) & R (\%) & F1 (\%) & IoU (\%) \\
\midrule
FCN~\cite{fcn} & 79.35 & 77.82 & 78.58 & 64.71 \\
SegNet~\cite{segnet} & 85.20 & 86.21 & 85.70 & 74.98 \\
Deeplabv3+~\cite{deeplab} & 89.24 & 90.91 & 90.07 & 81.93 \\
U-Net~\cite{unet} & 83.19 & 84.02 & 83.60 & 71.83 \\
PSPNet~\cite{psp} & 84.85 & 82.09 & 83.45 & 71.60 \\
HRNet~\cite{hrnet} & 86.77 & 85.92 & 86.34 & 75.97 \\
MA-FCN~\cite{mafcn} & 86.10 & 89.92 & 87.97 & 78.52 \\
Segformer~\cite{segformer} & 90.45 & 88.93 & 89.68 & 81.30 \\
TransUNet~\cite{transunet} & \textbf{93.82} & 89.33 & 91.52 & 84.37 \\
\midrule
STSUN & 93.07 & \textbf{91.20} & \textbf{92.13} & \textbf{85.40} \\
\bottomrule
\end{tabular}
\end{table}

In the context of bi-temporal binary change detection on the LEVIR-CD dataset, which features a large number of buildings of various sizes and styles undergoing changes, STSUN surpasses compared methods. Table \ref{levircd_result} shows that STSUN achieves the highest F1-score of 91.59\% and an IoU of 84.49\%. The strength of STSUN on this dataset lies in its robust temporal modeling capabilities, which allows for consistent feature representation across different time points. Furthermore, the LGWA mechanism's proficiency in extracting salient local changes while considering global context ensures high accuracy in detecting both small and large-scale building changes, minimizing missed detections and false alarms often encountered with heterogeneous scene elements.

\begin{table}[!ht]
\caption{Accuracy Comparison on the LEVIR-CD Dataset. The Best Values Are Highlighted in Bold.}
\label{levircd_result}
\centering
\begin{tabular}{lcccc}
\toprule
Methods & P (\%) & R (\%) & F1 (\%) & IoU (\%) \\
\midrule
FC-EF~\cite{fcef} & 86.91 & 80.17 & 83.40 & 71.53 \\
FC-Siam-Diff~\cite{fcef} & 89.53 & 83.31 & 86.31 & 75.91 \\
FC-Siam-Conc~\cite{fcef} & 91.99 & 76.77 & 83.69 & 71.96\\ 
DTCDSCN~\cite{dtcdstn} & 88.53 & 86.83 & 87.67 & 78.05\\ 
DSIFN~\cite{dsifn} & \textbf{94.02} & 82.93 & 88.13 & 78.77 \\
STANet~\cite{levircd_dataset} & 83.81 & \textbf{91.00} & 87.26 & 77.39 \\
SNUNet~\cite{snu} & 89.18 & 87.17 & 88.16 & 78.83 \\
HANet~\cite{hanet} & 91.21 & 89.36 & 90.28 & 82.27 \\
CDNet~\cite{cdnet} & 91.60 & 86.50 & 88.98 & 80.14 \\
DDCNN~\cite{ddcnn} & 91.85 & 88.69 & 90.24 & 82.22 \\
BIT~\cite{bit} & 89.24 & 89.37 & 89.30 & 80.68 \\
ChangeFormer~\cite{changeformer} & 92.05 & 88.80 & 90.40 & 82.47 \\
MTCNet~\cite{mtcnet} & 90.87 & 89.62 & 90.24 & 82.22 \\
MSCANet~\cite{mscanet} & 91.30 & 88.56 & 89.91 & 81.66 \\
AMTNet-50~\cite{amtnet} & 91.82 & 89.71 & 90.76 & 83.08 \\
\midrule
STSUN & 93.17 & 90.07 & \textbf{91.59} & \textbf{84.49} \\
\bottomrule
\end{tabular}
\end{table}

Finally, on the TSCD dataset, which presents a challenging multi-temporal binary change detection scenario with longer image sequences, STSUN achieves the best results, as indicated in Table \ref{tscd_result}, with an F1-score of 66.48\% and an IoU of 49.79\%. The TSCD dataset requires robust modeling of temporal dependencies across multiple observations. STSUN is specifically designed to handle inputs with varying temporal lengths and to model the continuous evolution of buildings, which enables STSUN to accurately identify changes in complex, evolving landscapes.

\begin{table}[!ht]
\caption{Accuracy Comparison on the TSCD Dataset. The Best Values Are Highlighted in Bold.}
\label{tscd_result}
\centering
\begin{tabular}{lcccc}
\toprule
Methods & F1 (\%) & IoU (\%) & OA (\%) \\
\midrule
FC-EF~\cite{fcef} & 53.39 & 37.86 & 97.03 \\
FC-Siam-Conc~\cite{fcef} & 41.51 & 28.05 & 96.60 \\
FC-Siam-Diff~\cite{fcef} & 39.65 & 26.83 & 96.48 \\
SNUNet-CD~\cite{snu} & 63.22 & 47.22 & 97.61 \\
USSFCNet~\cite{ultralightweight} & 55.68 & 39.80 & 97.30 \\
A2Net~\cite{lightweight} & 53.16 & 37.19 & 97.14 \\
SEIFNet~\cite{spatiotemporal} & 60.37 & 44.01 & 97.41 \\
Contrast-COUD~\cite{tscd_dataset} & 64.35 & 48.45 & 97.72 \\
TS-COUD~\cite{tscd_dataset} & 65.24 & 49.33 & 97.90 \\
\midrule
STSUN & \textbf{66.48} & \textbf{49.79} & \textbf{98.05} \\
\bottomrule
\end{tabular}
\end{table}

Figure \ref{Fig:building_result} presents inference results of STSUN from all four datasets. Visually, STSUN consistently produces accurate and complete segmentation and change maps. The results exhibit few false positives and false negatives, particularly in challenging areas such as those with small objects, intricate boundaries, or subtle temporal variations. For instance, in building segmentation on the WHU dataset (Figure \ref{Fig:building_result}(a)), STSUN generates sharp edges and complete building shapes. Similarly, for change detection tasks on WHU-CD (Figure \ref{Fig:building_result}(b)), LEVIR-CD (Figure \ref{Fig:building_result}(c)), and TSCD (Figure \ref{Fig:building_result}(d)), STSUN demonstrates superior capability in precisely localizing changed regions while maintaining the integrity of unchanged areas. This visual superiority can be attributed to the model's enhanced capability, derived from the synergistic operation of STSUN and LGWA, to learn highly discriminative features and effectively model contextual relationships across the spatial, temporal and spectral dimensions, resulting in outputs that are more coherent and closely aligned with ground truth.

\subsection{Overall Comparison on the LULC Scenario}

The efficacy of the proposed Spatial-Temporal-Spectral Unified Network is rigorously evaluated against SOTA methods on two challenging LULC datasets: LoveDA for single-temporal land cover classification and DynamicEarthNet for multi-temporal land cover classification.

On the LoveDA dataset, STSUN demonstrates superior performance, achieving the highest OA of 71.82\% and mIoU of 65.73\% as shown in Table \ref{loveda_result}. This leading performance can be attributed to STSUN's architecture, particularly the Local-Global Window Attention mechanism, which effectively captures both fine-grained local details and broader contextual information. This capability is crucial for accurately segmenting multiscale objects and navigating the complex scenes and backgrounds characteristic of the high-resolution LoveDA imagery.

\begin{table*}[!ht]
\caption{Accuracy Comparison on the LoveDA Dataset. The Best Values Are Highlighted in Bold.}
\label{loveda_result}
\centering
\begin{tabular}{lcccccccccc}
\hline
\multirow{2}{*}{Methods} & \multicolumn{7}{c}{F1-score per category (\%)}                       & \multirow{2}{*}{AF(\%)} & \multirow{2}{*}{OA(\%)} & \multirow{2}{*}{mIoU(\%)} \\ \cline{2-8}
                         & Background & Building & Road & Water & Barren & Forest & Agriculture &                     &                     &                       \\ \hline
U-Net \cite{unet} & 50.21 & 54.74 & 56.38 & 77.12 & 18.09 & 48.93 & 66.05 & 53.07 & 51.81 & 47.84 \\
DeepLabV3+ \cite{deeplab} & 52.29 & 54.99 & 57.16 & 77.96 & 16.11 & 48.18 & 67.79 & 53.50 & 52.30 & 47.62 \\
DANet \cite{danet} & 54.47 & 61.02 & 63.37 & 79.17 & 26.63 & 52.28 & 70.02 & 58.14 & 54.64 & 50.18 \\
ResUNet-a \cite{resunet_a} & 59.16 & 64.08 & 66.73 & 81.01 & 32.23 & 55.81 & 75.79 & 62.12 & 59.65 & 54.16 \\
DASSN \cite{dassn} & 57.95 & 66.90 & 68.63 & 76.64 & 44.35 & 54.96 & 70.49 & 62.85 & 60.35 & 55.42 \\
HCANet \cite{hcanet} & 66.39 & 70.76 & 75.11 & 88.29 & 51.14 & 63.92 & 81.07 & 70.95 & 69.47 & 62.77 \\
RAANet \cite{raanet} & 55.02 & 62.19 & 65.58 & 81.03 & 29.25 & 54.11 & 74.07 & 60.18 & 58.95 & 53.93 \\
SCAttNet \cite{scattnet} & 65.95 & 71.88 & 77.04 & 86.61 & 50.79 & 61.19 & 82.00 & 70.78 & 67.31 & 61.09 \\
A2FPN \cite{a2fpn} & 65.17 & 73.32 & 75.19 & 88.01 & 48.82 & 59.96 & 79.71 & 70.03 & 66.89 & 61.14 \\
LANet \cite{lanet} & 67.04 & 74.19 & 77.54 & 87.54 & 52.23 & 64.78 & 80.80 & 72.02 & 69.11 & 62.16 \\
MSAFNet \cite{msafnet} & 65.51 & 73.71 & 75.59 & 88.47 & 49.08 & 60.28 & 80.13 & 70.40 & 67.17 & 60.76 \\
CLCFormer \cite{clcformer} & 67.17 & 74.34 & 77.69 & 87.71 & 52.34 & 64.91 & 80.96 & 72.16 & 69.37 & 63.85 \\
SAPNet \cite{sapnet} & 67.50 & 75.06 & 78.12 & 88.35 & 53.10 & 65.50 & 81.30 & 73.04 & 70.12 & 63.45 \\
SSCBNet \cite{sscbnet} & 68.25 & 75.90 & \textbf{79.00} & 89.10 & 54.00 & 66.20 & 82.15 & 74.05 & 70.95 & 64.58 \\ \hline
STSUN & \textbf{68.72} & \textbf{76.13} & 78.83 & \textbf{90.21} & \textbf{54.73} & \textbf{67.03} & \textbf{82.67} & \textbf{74.81} & \textbf{71.82} & \textbf{65.73} \\
\hline
\end{tabular}
\end{table*}

For the multi-temporal DynamicEarthNet dataset, STSUN again surpasses existing methods, yielding the top scores across all reported metrics: SCS score of 29.9, BC of 38.9, and mIoU of 54.7 as shown in Table \ref{dynamicearthnet_result}. The success of STSUN on this dataset underscores the effectiveness of the STSUN in explicitly modeling the temporal dimension. By leveraging temporal metadata to ensure consistency across sequences of varying lengths, STSUN adeptly manages long-sequence image time series and discerns subtle land cover changes despite spectral variations across different time points—a key challenge in DynamicEarthNet.

\begin{table}[!ht]
\caption{Accuracy Comparison on the DynamicEarthnet Dataset. The Best Values Are Highlighted in Bold.}
\label{dynamicearthnet_result}
\centering
\begin{tabular}{lcccc}
\hline
Methods & SCS$\uparrow$ & BC$\uparrow$ & SC$\uparrow$ & mIoU(\%)$\uparrow$ \\ \hline
CAC \cite{cac} & 17.7 & 10.7 & \textbf{24.7} & 37.9 \\
U-Net \cite{unet} & 17.3 & 10.1 & 24.4 & 37.6 \\
TSViT \cite{tsvit} & 23.0 & 34.1 & 11.8 & 50.5 \\
UTAE \cite{utae} & 25.9 & 38.0 & 13.8 & 53.7 \\
A2Net \cite{a2net} & 22.2 & 32.9 & 11.5 & 47.2 \\
SCanNet \cite{scannet} & 24.8 & 35.8 & 13.9 & 53.0 \\
TSSCD \cite{tsscd} & 12.0 & 19.4 & 4.7 & 33.9 \\ \hline
STSUN & \textbf{30.3} & \textbf{38.9} & 21.8 & \textbf{55.7} \\
\hline
\end{tabular}
\end{table}

\begin{figure}[!ht]
 \centering
 	\includegraphics[width=\linewidth]{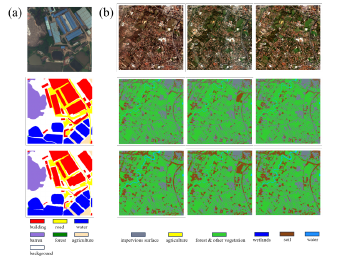}
 \caption{Sample inference results on for LULC scene datasets. The input images, ground truths and predictions are shown in the first, second and third rows, respectively. (a) LoveDA dataset sample. (b) DynamicEarthNet dataset sample.}
 \label{Fig:lulc_result}
\end{figure}

Figure \ref{Fig:lulc_result} presents the LULC semantic segmentation results of our proposed STSUN on the single-temporal LoveDA dataset (Figure \ref{Fig:lulc_result} a) and the multi-temporal Dynamic EarthNet dataset (Figure \ref{Fig:lulc_result} b). The method achieves strong performance across both, effectively navigating challenges such as the complex spectral-spatial features, multi-scale objects in LoveDA, and the inherent temporal variations within Dynamic EarthNet. This proficiency stems from STSUN’s ability to perform unified representation and modeling of remote sensing data across spatial, temporal and spectral dimensions. Specifically, the STSUN enables harmonized encoding and feature fusion across these dimensions, while the LGWA mechanism efficiently captures both local details and global contextual information, crucial for accurate LULC delineation in diverse scenarios.

\subsection{Ablation Study}

To ascertain the efficacy of the proposed unified strategy of STSUN and LGWA, ablation studies were performed on the TSCD dataset. The unified strategy of STSUN is denoted as "decoupled unification," which consistently preserves the independence of the temporal dimension throughout the unification process across various dimensions. As the temporal dimension is strongly correlated with dense prediction task types, this strategy enables joint modeling of multiple tasks on independent dimensions, learning the complementarity between each task. As comparative baselines, the decoupled unification strategy was replaced with a strategy involving the direct unification of spatial, temporal, and spectral dimensions at both the input and output stages (referred to as "coupled unification"). Specifically, in the input stage, this coupled unification strategy merges the temporal dimension with the channel dimension, without preserving an independent temporal dimension. The DUM is then used to map the combined channel dimension into a unified feature. In the output stage, the temporal dimension is separated from the channel dimension, and the DUM is employed to map the unified feature to the desired temporal length required by the task. Concurrently, LGWA was substituted with standard global attention. This experimental design serves to highlight the distinct advantages of STSUN in its independent extraction of temporal features and the capability of LGWA in the simultaneous capture of both local and global contextual information.

\begin{table}[!ht]
\caption{Ablation study on the TSCD Dataset. The Best Values Are Highlighted in Bold.}
\label{tab:ablation}
\centering
\begin{tabular}{cclll}
\hline
Strategy           & Attention                            & \multicolumn{1}{c}{F1(\%)}    & IoU(\%)   & OA(\%)    \\ \hline
Coupled Unification & \multicolumn{1}{l}{Global Attention} & 62.87                     & 45.85 & 94.38 \\
Coupled Unification & LGWA                                 & 63.22                     & 46.22 & 94.72 \\
Decoupled Unification              & \multicolumn{1}{l}{Global Attention} & 65.91                     & 49.15 & 97.21 \\
Decoupled Unification              & LGWA                                 & \multicolumn{1}{c}{\textbf{66.48}} & \textbf{49.79} & \textbf{98.05} \\ \hline
\end{tabular}
\end{table}

Table \ref{tab:ablation} displays the results of our ablation studies, demonstrating that both the decoupled unification strategy and LGWA outperform their respective baseline alternatives. Furthermore, their combined use leads to even greater performance improvements.

Specifically, the decoupled unification strategy enhances performance by preserving an independent temporal dimension. This allows STSUN to extract features from remote sensing images at various time steps during the encoder stage. Subsequently, in the feature fusion stage, feature-level fusion effectively reduces interference from irrelevant information \cite{fccdn}. Moreover, when performing specific tasks, this strategy enables more effective integration with task embeddings and allows for the adjustment of temporal length to suit task requirements. This capability facilitates the joint modeling of multiple dense prediction tasks, leveraging data from various tasks to collectively improve their individual performance.


In a similar vein, LGWA offers advantages over standard global self-attention. By employing a combination of variously shaped local windows alongside global self-attention mechanisms, LGWA is capable of concurrently extracting a rich tapestry of local features and comprehensive global context. This simultaneous extraction of multi-level features is crucial for enabling the model to effectively perform dense prediction tasks across various scales.

\subsection{Unification of Dense Prediction Tasks}



Semantic segmentation, binary change detection, and semantic change detection are prevalent dense prediction tasks in remote sensing, exhibiting significant similarities and complementarities. To validate the capability of STSUN to unify these diverse dense prediction tasks, we conducted a comparison experiment on four datasets in the building scenario. Specifically, we compare the performance of STSUN models trained individually on each of the four datasets against a single STSUN model trained jointly on all datasets. For a given dataset, the former is denoted as $STSUN_{\{dataset\}\_single}$ and the latter as $STSUN_{\{dataset\}\_unified}$.

The results, presented in Table \ref{task_unification}, demonstrate that $STSUN_{\{dataset\}\_unified}$ consistently outperforms its single-task counterparts across all building datasets. This performance gain is attributed to two factors. First, joint training on data from multiple dense prediction tasks exposes the model to a richer and more diverse set of samples, enhancing its ability to learn comprehensive spatial-temporal-spectral features from remote sensing imagery. Second, the model capitalizes on the complementary nature of these tasks. This synergy, analogous to a multi-task learning paradigm, enables the learning of more powerful and robust feature representations, thereby elevating the model's performance across all individual tasks.

\begin{table}[!ht]
\caption{Comparison over Task Unification on Four Building Scenario Datasets.  The Best Values Are Highlighted in Bold.}
\label{task_unification}
\centering
\begin{tabular}{lcccc}
\toprule
Methods & P (\%) & R (\%) & F1 (\%) & IoU (\%) \\
\midrule
$STSUN_{whu\_single}$ & 95.24 & 94.30 & 94.77 & 90.06 \\
$STSUN_{whu\_unified}$ & \textbf{95.71} & \textbf{94.87} & \textbf{95.29} & \textbf{91.00} \\
\midrule
$STSUN_{whucd\_single}$ & 92.22 & 90.89 & 91.55 & 84.42 \\
$STSUN_{whucd\_unified}$ & \textbf{93.07} & \textbf{91.20} & \textbf{92.13} & \textbf{85.40} \\
\midrule
$STSUN_{levircd\_single}$ & 92.94 & 89.63 & 91.25 & 83.92 \\
$STSUN_{levircd\_unified}$ & \textbf{93.17} & \textbf{90.07} & \textbf{91.59} & \textbf{84.49} \\
\midrule
$STSUN_{tscd\_single}$ & 66.91 & 65.08 & 65.98 & 49.23 \\
$STSUN_{tscd\_unified}$ & \textbf{67.32} & \textbf{65.67} & \textbf{66.48} & \textbf{49.79} \\

\bottomrule
\end{tabular}
\end{table}

\subsection{Dense Prediction with Flexible Semantic Category Set}



Remote sensing scenarios often encompass distinct sets of ground objects, necessitating different sets of semantic categories. For instance, the two LULC datasets employed in this study exhibit this variance: the LoveDA dataset uses the semantic set \{\texttt{building}, \texttt{road}, \texttt{water}, \texttt{barren}, \texttt{forest}, \texttt{agriculture}, \texttt{background}\}, whereas the DynamicEarthNet dataset uses \{\texttt{impervious surface}, \texttt{agriculture}, \texttt{forest}, \texttt{wetlands}, \texttt{soil}, \texttt{water}\}. To demonstrate STSUN's ability to handle flexible semantic category sets for dense prediction, we designed an experiment using these two datasets. For this experiment, the DynamicEarthNet dataset was used exclusively for semantic segmentation, utilizing only the image pairs with semantic annotations. We compare four model configurations:

\begin{enumerate}

\item $STSUN_{loveda}$, trained exclusively on the LoveDA dataset to predict its 7-category set.

\item $STSUN_{dynamic.}$, trained exclusively on the DynamicEarthNet dataset to predict its 6-category set.

\item $STSUN_{fixed}$, trained jointly on both datasets to predict a fixed, 10-category set corresponding to the union of their individual category sets.

\item $STSUN_{flexible}$, trained jointly on both datasets but dynamically predicting from the appropriate category subset for each respective dataset.

\end{enumerate}

As shown in Table \ref{class_unification}, $STSUN_{flexible}$ outperforms $STSUN_{fixed}$ on both datasets, while achieving performance comparable to the specialist models, $STSUN_{loveda}$ and $STSUN_{dynamic.}$. The degraded performance of $STSUN_{fixed}$ is expected. This model is constrained to predict over the union of all categories, even for an image from a scene that does not contain certain categories. Forcing a single, fixed-size output space across scenes with disparate semantic sets introduces ambiguity and negatively impacts performance on each respective scene. In contrast, $STSUN_{flexible}$ dynamically adapts its predictive output to the relevant subset of semantic categories for each scene, which explains why its performance is on par with the individually trained models. Crucially, however, $STSUN_{flexible}$ offers superior model efficiency, as a single, unified model can be deployed across scenes with different semantic category sets without requiring any additional training or fine-tuning.

\begin{table}[!ht]
\caption{Comparison over Category Unification on Two LULC Scenario Datasets. The Best Values Are Highlighted in Bold.}
\label{class_unification}
\centering
\begin{tabular}{lcccc}
\hline
\multicolumn{2}{l}{$Model_{loveda}$} & AF(\%) & OA(\%) & mIoU(\%) \\ \hline
\multicolumn{2}{l}{$STSUN_{loveda}$} & 74.66      & 71.70      & 65.59        \\
\multicolumn{2}{l}{$STSUN_{fixed}$} & 74.03      & 70.49      & 64.67        \\
\multicolumn{2}{l}{$STSUN_{flexible}$} & \textbf{74.81} & \textbf{71.82} & \textbf{65.73}        \\ \hline
$Model_{dynamic.}$        & SCS        & BC     & SC     & mIoU(\%) \\ \hline
$STSUN_{dynamic.}$        &   30         &   38.4     &   21.6     &    55.3      \\
$STSUN_{fixed}$        &      29.4      & 38.7 &   20.1     &    54.2      \\
$STSUN_{flexible}$        & \textbf{30.3} & \textbf{38.9} & \textbf{21.8} & \textbf{55.7}          \\ \hline
\end{tabular}
\end{table}

%
%
\section{DISCUSSION}

\subsection{Unified Input and Output Modeling}

A core strength of the proposed STSUN framework lies in its capacity to accommodate arbitrary input and output configurations across the spatial, temporal, and spectral dimensions, overcoming the rigidity inherent in conventional architectures. This adaptability is particularly advantageous for processing the heterogeneous data prevalent in remote sensing applications. 

For input data, the spectral dimension $C_1$ is strongly correlated with sensor modality. The flexibility of STSUN in spectral dimension allows it to seamlessly ingest and model data from diverse sources, including RGB, Synthetic Aperture Radar, multispectral, and hyperspectral sensors, within a singular and unified model. At the same time, the temporal dimension $T_1$ is directly related to the time span and acquisition frequency of an image series. The ability to handle variable temporal lengths enables STSUN to process datasets with disparate temporal characteristics without modification. Moreover, the spatial dimensions $H_1, W_1$ are tied to the geographic coverage and spatial resolution of the imagery. STSUN's architectural design is agnostic to input image size, thus capably handling data from different regions coverage and spatial resolutions.

This principle of unification extends symmetrically to the model's output, enabling versatile and efficient inference. The configuration of the output temporal dimension $T_2$ is intrinsically linked to the dense prediction task being executed. This allows STSUN to unify semantic segmentation, binary change detection, and semantic change detection, facilitating joint modeling that can exploit complementary regularities between these tasks to enhance overall performance. Furthermore, the output channel dimension $C_2$ corresponds directly to the set of predicted semantic categories. By treating the category set as a flexible parameter, STSUN can adapt to diverse prediction class sets across various remote sensing scenarios. This obviates the need for extensive retraining or the maintenance of multiple specialized models, representing a significant step towards more scalable and universally applicable frameworks for remote sensing data analysis.

\subsection{Unified Multitask Learning}

STSUN introduces a unified framework for dense prediction, capable of concurrently addressing semantic segmentation, binary change detection, and semantic change detection within a single model. This unification is principally achieved through the trainable task embeddings and the Dimension Unification Module. This approach aligns with the core tenets of multi-task learning (MTL), which posit that jointly learning related tasks can lead to improved generalization by leveraging shared representations \cite{multitask}. By training on a comprehensive dataset aggregated from these distinct tasks, our model develops more robust features than models trained in a single-task paradigm \cite{mtl_overview}.

Moreover, our framework marks a departure from conventional MTL applications in the remote sensing domain. Prevailing methods typically depend on the availability of paired datasets, where each geographical location has labels for all tasks, and commonly employ multiple task-specific decoder heads to generate predictions. In stark contrast, the key strengths of our model are its capacity to be trained on non-paired data from disparate tasks and its reliance on a single, shared decoder head for all predictions.

The implications of this design are twofold. First, the ability to utilize non-paired data dramatically expands the pool of usable training imagery, addressing the data scarcity issue in the field. Second, the single shared decoder enforces the learning of a more cohesive and generalized feature representation, promoting more effective knowledge transfer between tasks. This results in a framework that is not only more data-efficient but also exhibits enhanced adaptability and robustness across a diverse range of dense prediction challenges.

\subsection{Limitations and Expectations}

While the proposed Spatial-Temporal-Spectral Unified Network demonstrates considerable promise in harmonizing the analysis of heterogeneous remote sensing data, its current instantiation presents certain limitations which, in turn, illuminate clear and compelling trajectories for future research.

First, the operational flexibility of our framework is presently contingent upon the provision of explicit metadata at inference time, specifically in the form of predefined task and category embeddings. This requirement, while effective, curtails the model's autonomy and presupposes a level of a priori knowledge about the analytical objective. A significant advancement would be to imbue the model with the capacity to implicitly infer the task and desired output configuration directly from the contextual cues within the input data stream. Future work could explore methodologies grounded in meta-learning or employ sophisticated attention mechanisms that learn to dynamically weigh different aspects of the data, thereby deducing the analytical intent without explicit instruction.

Second, while our method achieves a critical unification at the data format and architectural input level across the spatial, temporal, and spectral dimensions, this does not inherently guarantee robust semantic generalization across the vast heterogeneity of remote sensing scenes, sensor modalities, and non-training semantic category sets. A promising path to surmount this limitation lies in elevating the STSUN framework into a large-scale, foundational model for Earth observation by employing advanced self-supervised or multi-modal pre-training strategies. The objective would be to produce a model that captures the fundamental structures of spatial-temporal-spectral data, enabling highly transferable representations that dramatically improve performance on a wide array of tasks with minimal, or even zero-shot, fine-tuning.

Finally, the potential of our unified modeling paradigm has thus far been demonstrated exclusively within a supervised learning context. A significant opportunity exists to extend this framework into the self-supervised and vision-language domains. By pre-training the unified vision model on enormous, unlabeled remote sensing archives, one could construct a powerful Remote Sensing Vision Foundation Model. Such a model, pre-trained to comprehend the elemental structure of diverse STS data, could then be adapted with remarkable data efficiency for specialized downstream tasks. Concurrently, applying this unified input approach to Remote Sensing Vision-Language Models holds transformative potential. It would permit the training of these models on vast corpora of image-text pairs without the need for cumbersome, modality-specific engineering to handle varying spectral or temporal dimensions in the imagery. This would not only streamline the development and enhance the performance of RS-VLMs but also make them natively adaptable to the full diversity of remote sensing data, fostering a new generation of models capable of nuanced, cross-modal understanding of our planet.

%
%
\section{CONCLUSION}


This study introduced the Spatial-Temporal-Spectral Unified Network, a novel architecture designed to overcome the critical limitations of model rigidity in remote sensing dense prediction. By establishing a unified representation for arbitrary spatial, temporal, and spectral data configurations, STSUN demonstrates exceptional adaptability to heterogeneous datasets. STSUN uniquely integrates semantic segmentation, binary change detection, and semantic change detection within a single model via trainable task embeddings, obviating the need for specialized, task-specific architectures. Moreover, STSUN utilizes trainable semantic category embeddings to perform dense prediction for multiple prediction class setting without requiring model retraining. Comprehensive experimental validation on multiple datasets confirmed that a single STSUN model consistently adapts to varied data inputs and outputs, unifing multiple dense prediction tasks and prediction category settings, achieving or exceeding state-of-the-art performance. We envision STSUN to serve a baseline for universal remote sensing dense prediction models, mitigating the need for task-specific designs and extensive retraining.



%
%

\bibliographystyle{IEEEtran}

\bibliography{reference}

\vfill
\end{document}